\documentclass{article}

\usepackage{arxiv}

\usepackage[utf8]{inputenc} 
\usepackage[T1]{fontenc}    
\usepackage{hyperref}       
\usepackage{amsmath,amssymb,amsfonts}
\usepackage{algorithmic}
\usepackage{graphicx}
\usepackage{graphbox}
\usepackage{subcaption}
\usepackage[table]{xcolor}
\usepackage{textcomp}
\usepackage{url}

\newcommand{\figureref}[1]{Figure~{\ref{#1}}}

\newcommand{\eqnref}[1]{Eqn.~(\ref{#1})}
\newcommand{\tableref}[1]{Table~{\ref{#1}}}
\newcommand{\sectionref}[1]{Section~{\ref{#1}}}
\newcommand{\snip}[4]{
	\begin{minipage}{#1}
		\centering
		\renewcommand{\tabcolsep}{1pt}
		\begin{tabular}{cc}
			\includegraphics[width=0.49\linewidth]{#2} & \includegraphics[width=0.49\linewidth]{#3} \\
			\multicolumn{2}{c}{
				\begin{minipage}{0.98\linewidth}
					\centering
					{#4}
				\end{minipage}
			}
		\end{tabular}					
	\end{minipage}
}

\title{SummaryNet: A Multi-Stage Deep Learning Model for Automatic Video Summarisation}

\author{
 Ziyad Jappie \\
  School of Computer Science and Applied Mathematics\\
  University of the Witwatersrand\\
  Johannesburg, South Africa \\
  \texttt{ziyadjappie@gmail.com} \\
   \And
 David Torpey \\
  School of Computer Science and Applied Mathematics\\
  University of the Witwatersrand\\
  Johannesburg, South Africa \\
  \texttt{torpey.david93@gmail.com} \\
  \And
 Turgay Celik \\
  School of Computer Science and Applied Mathematics\\
  University of the Witwatersrand\\
  Johannesburg, South Africa \\
  \texttt{celikturgay@gmail.com} \\
}

\begin{document}
\maketitle
\begin{abstract}
Video summarisation can be posed as the task of extracting important parts of a video in order to create an informative summary of what occurred in the video. In this paper we introduce SummaryNet as a supervised learning framework for automated video summarisation. SummaryNet employs a two-stream convolutional network to learn spatial (appearance) and temporal (motion) representations. It utilizes an encoder-decoder model to extract the most salient features from the learned video representations. Lastly, it uses a sigmoid regression network with bidirectional long short-term memory cells to predict the probability of a frame being a summary frame. Experimental results on benchmark datasets show that the proposed method achieves comparable or significantly better results than the state-of-the-art video summarisation methods.
\end{abstract}


\section{Introduction}
\label{sec:introduction}

Video summarisation has recently become an active area of research due to a myriad of possible applications, such as in the entertainment industry, sports, and surveillance \cite{Yang2015, Yuan2019}. There are two video summarisation problems. The first is the dynamic variant (the one which we attempt to solve in this work), which is defined as finding a set of short video snippets, which when combined together, create the video summary. The second is static video summarisation, in which key images are taken from the video in order to create the summary. It can be argued that videos are currently the most important source of visual data \cite{Rochan2018}. Solving this problem will likely have a large impact in many domains that have some form of video records. Both audio and visual inputs can be used for summarisation \cite{Takahashi2018}. Users would ideally like to browse through videos quickly to get an idea of the content. This will enable faster browsing of large video datasets, as well as better grouping and access to these videos.

Deep learning, and in particular, convolutional neural networks (CNNs), have shown to be very useful in many areas of computer vision. Such domains include image recognition \cite{He2016,Krizhevsky2017}, image generation \cite{Isola2017}, and most importantly, video recognition and modeling \cite{Simonyan2014}. Bearing these advances in mind, we opt to leverage a recent three-dimensional CNN architecture I3D \cite{Carreira2017} that is the current state-of-the-art architecture for action recognition. I3D consists of two independent network streams - one processing RGB videos, and the other processing optical flow videos. We utilise these networks to extract high-level feature representations for our videos. These representations are then condensed using an encoder-decoder architecture in order to learn a richer, compressed, noise-free representation of the videos. This compressed representation is used to predict the probability that a particular frame of the video is a summary frame.

We frame our approach to identifying summary regions of a video as a supervised learning problem which takes videos and corresponding human labeled summaries as training data to explicitly learn the way humans summarise videos. Our contributions are applying the state-of-the-art action recognition CNN model to video summarisation by separately modeling the appearance and motion information of a video using the two streams of the I3D network. Previous approaches typically model only the appearance information (using either 2D or 3D CNNs), however, some recent approaches have adopted a separate stream for modelling the temporal dynamics as well. In this work we explicitly model the temporal component of the videos using I3D's optical flow-based CNN stream. This, we posit, allows for a much richer representation of a video. Further contributions include introducing an encoder-decoder architecture to learn a better representation of the videos from the I3D features. This allows a model to more accurately estimate the probability of a particular frame being a summary frame. To the best of our knowledge, the proposed method is first end-to-end learning architecture which uses optical flow for the temporal modeling and the features from the latent space learned via  encoder-decoder model to make frame level predictions for video summarisation. 

The remainder of the paper is organised as follows. \sectionref{sec:related_work} provides a literature survey on related work. The proposed SummaryNet method is described in detail in \sectionref{sec:Method}. Experimental results are given in \sectionref{sec:Experimental Results} and, finally, the paper is concluded in \sectionref{sec:Conclusion}. 

\section{Related Work}
\label{sec:related_work}

\subsection{Conventional Approaches}

Earlier approaches to video summarisation typically employ hand-crafted feature extraction, followed by feeding these into an algorithm for further processing \cite{Potapov2014, Gygli2014, Song2015}. The most common, widely-used algorithm for video segmentation in the video summarisation domain is the Kernel Temporal Segmentation (KTS) algorithm \cite{Potapov2014}. The KTS algorithm aims to group a video into semantically-consistent segments, formally defined as temporal volumes. The original video of temporal duration $n$ is broken up into $m$ non-uniform snippets of size $l_i$, $i=0,...m$, and $l_i < n$. To get shot boundaries, one would generally compare successive frames, relying on image descriptors. However, KTS compares all pairs of frames which allows it not only to produce shot boundaries, but also general change points within the video. These change points are identified by detecting `jumps' in the video input signals. Each video frame is represented by a feature vector of dimensionality $k$.

Other early approaches such as \cite{mayufei} model user attention through a set of audio-visual model features. Each frame of a video gets assigned an attention value. Audio, visual, and linguistic features are extracted and modelled. A fusion scheme for these various features by way of a linear combination is employed. A user-attention curve is then created from this fusion of features, and the peaks of this curve are selected as the summary frames. \cite{doulamis} create a fuzzy representation of a video in the form of a multidimensional fuzzy histogram. Colour and motion information is extracted on a per-frame basis, and classified using the aforementioned fuzzy representation. A scheme is also employed for removal of redundant frames. \cite{pitas} create a video segmentation method that is able to detect cuts, as well as fade-ins and fade-outs. This is done by computing similarities between frames. If the standard deviation of the similarities within a particular segment of the video is low, any of the frames in the segment can be selected as a summary frame. If the standard deviation is high, the frames are clustered until the standard deviation is below a threshold. Lastly, \cite{mundur} use Delaunay Triangulation to cluster the frames of a video. This proposed algorithm removes visual-content redundancy among the frames. This triangulation process results in semantically-similar frames being grouped together. In order to perform the clustering, frames are represented as multi-dimensional data points.

\subsection{Deep Learning Approaches}

Recently, the most popular approach to tackle the video summarisation problem is to first extract deep feature representations for each frame, and subsequently feed them into a model to learn how to extract video summaries. \cite{Zhang2016, Ji2019, Mahasseni2017, Zhou2018} all extract GoogLeNet features, which are spatial, frame-level feature representations. Image-based deep features \cite{Krizhevsky2017, He2016, Simonyan2015, Szegedy2015, Zeiler2014} are not generally suitable for videos due to their lack of motion modelling \cite{Tran2014}. This brings rise to the need for a generic video descriptor that helps solve large scale video tasks. One of the first models to tackle this was \cite{Tran2014}. We instead opt for the more modern I3D \cite{Carreira2017} architecture, due to its explicit modelling of motion (temporal) information through the optical flow stream. This architecture has shown very promising results for video processing in particular for action recognition. It consists of two separate networks: one for optical flow videos \cite{Tu2019}, and another for the raw RGB videos. These types of models are typically referred to as two-stream networks \cite{Simonyan2014}. Once these two models are trained, a novel video is classified by averaging the two networks' outputs. The advantage of this I3D architecture is that it uses both spatial and temporal information to model the video.

Previous methods often utilise long short-term memory (LSTM) networks to model the temporal dependencies of video summarisation. As such, \cite{Yang2015} employ LSTM cells in their Robust Recurrent autoencoder (RRAE) model. They also use a bidirectional RNN which, as the name suggests, flows both forward and backward through the network. This enables all information to be captured once the model is trained--the model is trained on \textit{inputs from the past and the future} \cite{Goodfellow2017}. Lastly, they extract C3D features as inputs into their network.

In \cite{Zhang2016}, they first extract GoogLeNet features for each frame. A bidirectional LSTM model is then trained on these features. However, a variation of the LSTM model (vsLSTM) is introduced, which combines the LSTM layers' hidden states and visual features with a multi-layer perceptron. They then extend this to another variation called dppLSTM (determinantal point process LSTM), which aims to increase the diversity in the selected frames by eliminating redundant frames.

Usually supervised approaches achieve better performance than unsupervised approaches \cite{Ji2019}. However, in \cite{Ji2019} they opt for the more challenging  unsupervised approach to solve the video summarisation problem in the hope to develop a more robust system. Their key idea was to develop a network with an attention mechanism to mimic human summaries. To achieve this, they extract GoogLeNet features for each frame. They then create an attentive encoder-decoder network for Video Summarisation (AVS). The encoder is constructed using a bidirectional LSTM to encode contextual information amongst the video inputs, similar to \cite{Zhang2016}. The decoder uses an attention-based LSTM network, allowing it to focus on only a subset of inputs by increasing attention weights. By adopting this method, they achieved state-of-the-art results. Another unsupervised approach is employed by \cite{Mahasseni2017}, who uses a generative adversarial network for their summary predictions. It consists of a summariser network and a discriminator network. The summariser is an auto-encoder LSTM network, whereas the discriminator is another LSTM model aimed at distinguishing the original video from the reconstructed video. The reason they decided to use a generative approach is because specifying a suitable distance metric between deep features is very challenging. They use GoogLeNet feature representations as inputs to their generative adversarial network. Similarly, \cite{Yuan2019} also used an AE in order to extract the most salient spatial features of the video input. They used two AEs: one for the input of the video sequence and one for the meta-data information. Thereafter, they correlate the two AEs' latent subspaces. One of the most unique methods for solving this problem was done by \cite{Zhou2018}. They proposed a reinforcement-learning-based network, where they designed their own reward function which judges how diverse and representative the generated summaries are. Here they use a encoder-decoder model where the encoder is a CNN architecture used as a feature extractor, namely GoogLeNet. For the decoder, they employ an RNN with an LSTM cell to capture temporal information. Their diversity reward measures how dissimilar two frames are in the feature space, while the representative reward measures how well the generated summary represents the original video.

\cite{Rochan2018} proposed a fully-convolutional sequence network (FCSN) to solve this problem. Their key contribution is to draw similarities between video summarisation and semantic segmentation, which is essentially assigning a class label to each pixel in an image. FCSN consists of a stack of convolutions whose size grows as the network gets deeper. This is aimed to capture long-range dependencies among frames. The architecture used in FCSN is an encoder-decoder formulation. The encoder network is tasked with processing input frames features to capture both high level semantic features and to capture long-range structural relationships between the input frames. The decoder is tasked with producing a sequence of 0/1 class labels. The design of this network was inspired by existing models available in semantic segmentation. They sub-sample the input video at 2 fps and extract GoogLeNet features as frame-level feature descriptors for each frame in the video.

\cite{CSNet} introduced a unsupervised model for discriminative feature learning. Here, they introduce a regularisation loss term to address the temporal dependency in LSTM based methods. They then design a novel two-stream network called Chunk and Stride Network (CSNet). These two newly created feature sequences are passed through a LSTM model and later merged back to get a final score. They also introduce attention mechanisms to handle dynamic information of videos by using the CNN based features for each frame. Another interesting unsupervised technique that follows \cite{Mahasseni2017} closely was presented by \cite{CycleNet}. The system is a two-step approach. The first being a selector network that simply gets the summary videos via frame-level values. This is optimised by a second network that is a cycle-consistent adversarial LSTM evaluator. This is used to evaluate the quality of the summary by maximising the mutual information between the summary video and original video. This is has a GAN-VAE structure that discriminates summary videos from original videos. They call this the Cycle-SUM network.

One of the first two-stream networks for first person videos was introduced by \cite{Yao2016}. They call their network two-stream Deep Convolution Neural Networks (DCNN). One stream is to address the appearance information, while the other models the temporal dynamics across the video's frames. They extract AlexNet features for the appearance information and C3D features \cite{Tran2014} for the temporal dynamics. These two networks are combined using late fusion. The DCNN system is evaluated on a first person dataset they created using mean average precision (mAP).

In this brief summary we see that previous methods typically extract spatial features only. Some of these are in the form of 2D spatial features (ResNet, GoogLeNet, and AlexNet), while others are 3D spatial features (C3D). C3D itself has inherently limited modelling of the temporal dynamics of a video. Other previous approaches dealt with temporal information in the architecture by introducing some form of recurrence (LSTM). In this work, we model both the spatial (appearance) and temporal (modtion) dependencies inherent in the videos through the use of the two independent I3D CNN streams.

\section{Proposed SummaryNet Method}
\label{sec:Method}
Our method consists of the following high-level steps:
\begin{enumerate}
    \item Fine-tuning both network streams of I3D;
    \item Obtaining high-level feature representations for both appearance and motion;
    \item Obtaining compressed I3D feature representations using encoder-decoder network;
    \item Training a recurrent regression model to estimate probability of a frame being a summary frame;
    \item Using the KTS and 0-1 knapsack algorithms for selecting summary snippets, and evaluation.
\end{enumerate}
Firstly, we fine-tune the weights of the I3D networks by training on the benchmark datasets. We employ transfer learning by using I3D networks pre-trained on the large-scale Kinetics dataset \cite{kinetics_download}. Thereafter, we are able to extract feature representations for each frame in the form of one vector for the RGB stream, and one for the optical flow stream. Next, we learn a compressed representation of these aforementioned RGB and optical flow representations that contain only the most salient features. This is achieved through the use of a encoder-decoder network. This compressed representation is used to estimate the probability of a frame being a summary frame. These probabilities are then used to compute logical contiguous sequences of frames (keyshots) that correspond to potential summaries of the video.

\subsection{Pre-processing}
\label{sec:pre_procesing}
The video files (typically in MPEG-4 or MP4 format) are read into the system memory as RGB images for each frame. We then resize each frame to $n \times m \times 3$ which corresponds to the height, width, and number of channels for each frame, respectively. While resizing, we preserve the aspect ratio of the video. That is, the smallest side between the width and the height is resized to 128 pixels, and the other size is resized according to the aspect ratio. Preserving the aspect ratio results in more realistic optical flow estimates, which should result in richer optical flow-based features.

We then randomly extract crops of size $16 \times 112 \times 112 \times 3$, which correspond to snippets of temporal resolution $16$ (i.e. snippets of $16$ frames in length). A step size of 1 frame is used, and the target for each cube is the target associated with the middle frame of the cube. For example, if we have a cube of length $16$, the target that is assigned to the frame at position $8$ is the target for the entire cube. The aforementioned process is also used for the optical flow videos. The optical flow videos are normalised using min-max normalisation, since there is no predefined finite range for optical flow values.

\subsection{Pre-trained models/features}
In this paper we compare using ResNet features with using I3D features. ResNet features will serve as the baseline model. By using I3D features we aim to show the importance of incorporating both appearance (spatial) and motion (temporal) features in order to improve the overall performance of the system.

\subsubsection{ResNet Features}
The model weights are initialised with the ImageNet weights available in the Keras library. We fine-tune the model by adding our own classification layer and removing the existing one. Once the model is fine-tuned on a sample of our data, we take the outputs of the second-last layer in this network as a feature representation of each individual frame. These feature vectors now represent each associated frame. Thus, we can define a mapping for this process as $f: \mathbb{R}^{m \times n} \mapsto \mathbb{R}^{j}$, where $f$ is a function mapping the image frame $I \in \mathbb{R}^{m \times n}$ of dimension $m \times n$ to a vector $\mathbf{v} \in \mathbb{R}^{j}$, where $j$ refers to the number of neurons in the second-last layer of the ResNet network.

\subsubsection{I3D features}
Similarly to ResNet, this model is also fine-tuned and used as a feature extractor. This architecture consists of two streams, as mentioned in Section \ref{sec:related_work}, namely the RGB stream and the dense optical flow stream. This architecture is fine-tuned on samples of our data and then used as a feature extractor to get a high-level, semantic feature representation of our data.

Since I3D features are the basis of how we intend to incorporate temporal information into our system, we will delve deeper into this. Once the cubes are created, as mentioned in Section \ref{sec:pre_procesing}, we feed our data into the I3D network, which consists of two separate networks: one for the RGB data, and one for the optical flow data. Thus, we fine-tune these two networks by removing the previous classification layer and adding our own sigmoid layer for classification. Each snippet is given the value of 1 or 0 depending on whether it is a summary or not. The weights used as a starting point on the I3D models are the weights from pre-training on the ImageNet and Kinetics datasets. These weights are available online \cite{kinetics_download}.

We then feed all our videos through these networks to get feature representations for each frame. This process involves creating cubes as mentioned above with a stride of 1 frame. We pass all these through the network and take the features in the second last layer of the network which is of size $1024$. This vector serves as a feature representation for the middle frame of the cubes we created. Formally, given a video $V \in \mathbb{R}^{N \times 112 \times 112 \times 3} \mapsto \mathbb{R}^{M \times 1024}$, where $N$ represents the number of frames in a video, and $M$ is the resulting number of features, $M < N$ since we apply no wrapper for the edges of the video. Furthermore, we can see the mapping for each individual cube as, $g : \mathbb{R}^{n \times 112 \times 112 \times 3} \mapsto \mathbb{R}^{1 \times 1024}$, where $n$ is the size of each cube. Note that this process is done for both the RGB stream and the optical flow stream.

These two models above (RGB and optical flow) are trained for 8 epochs each and the best model parameters are used. This is selected by which epoch returns the lowest loss value for training and these are taken as the optimal parameters of the network.

\subsection{SummaryNet}
\label{sec:ResearchMethod}

\begin{figure}
	\centering
	\includegraphics[width=0.99\linewidth]{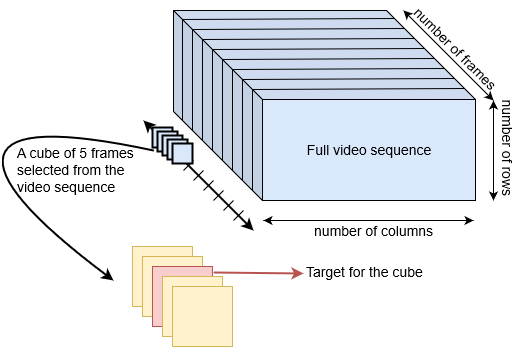}
	\caption{The target associated with the middle frame of the current video cube represents the target for the entire cube.}
	\label{fig:preprocessing}
\end{figure}

The approach we have decided on pursuing is two-fold. We train a bespoke \textbf{encoder-decoder} model offline (phase 1), whose latent feature representations are used to train a recurrent regression model (phase 2). This involves dividing the full bank of features into smaller feature snippets and feeding it through the model, similar to the process described in \figureref{fig:preprocessing}. The encoder part consists of 3 hidden layers: 2 densely connected and the penultimate one convolutional. The decoder leverages 2 fully-connected hidden layers. After the encoder-decoder model is trained, we use the encoder part as a feature extractor for each video snippet which is of size $n \times 512$, where $n$ is the length of the cube. We then feed these feature representations, or code layers, into a newly-trained recurrent regression model. The full pipeline, from I3D to the regression model, is known as the SummaryNet model.

\begin{figure*}
	\centering
	\includegraphics[width=0.99\linewidth]{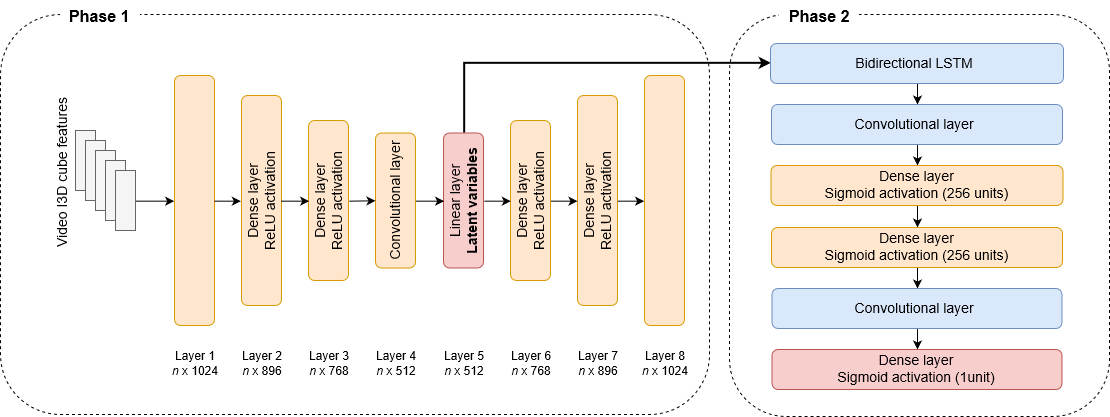}
	\caption{SummaryNet model is two-fold. Firstly, an encoder-decoder model is trained and the latent variables are extracted. Secondly, the latent variables are used as features to the second SummaryNet model. The second model is responsible for predicting the frame-level scores. The values for the number of neurons per layer in the networks were found using grid search.}
	\label{fig:model_arch}
\end{figure*}

This model consists of 5 hidden layers. The first of these hidden layers is a LSTM layer which aims to model long-term temporal dependencies. The second and fifth layers are convolutional. Hidden layers 3 and 4 consist of multi-layer perceptrons (MLPs), each with 256 units and sigmoid activations, as per \cite{Zhang2016} (apart from the final layer which is 1 sigmoidal unit). This is used to classify whether a snippet is part of the summary or not. The full flow of the SummaryNetw architecture can be seen in \figureref{fig:model_arch}.

\subsection{Training}
Since the datasets have multiple user annotations, we process this data to give us a single ground truth vector of size $1 \times N$, where $N$ is the number of frames per video. Along with the multiple user annotations, we are supplied with single frame-level scores for each video, which is the average score of all users. We formulate this task as a binary classification problem: if a certain number, $x$, users regard frame $m$ as a summary, then we give the value of $1$, else the value is $0$. We call this process `binarising' the video targets.

For each dataset, there is a split between training and testing which consists of a 80$\%$ training and 20$\%$ testing. Back-propagation with the Adam optimiser is used with the default Keras settings to minimize the loss function with a learning rate of $\theta$ and a batch size of 8. Since we have two classes, we will be using binary cross-entropy as our loss function. When training, we sub-sample 2fps as done in \cite{Zhang2016}, since the scenes are slowly changing in the datasets considered. The training process is run for 8 epochs, and the best model based on smallest validation loss is selected as the final model.

\subsection{Testing and Evaluation Metrics}
The system is tested to evaluate the summary results produced by the model. The evaluation is performed using a 5-fold cross validation framework. The testing of the system has been constructed to follow \cite{Zhang2016} as close as possible. 

To test the system, we feed in both the RGB and optical flow feature representations. We create the snippets of 5 frames in length, i.e. we end up with snippets of size ${5 \times 1024}$. The model predicts the value for the middle frame of the snippet, we also take a stride of 1 frame when creating these snippets so that we may get predictions for each individual frame. Once the model predicts the frame-level scores, we do some post-processing. Finally, it is evaluated against previous results. 

\subsubsection{Post-processing}
Our model makes frame-level probability estimates. We use these predictions to extract the most summary-like, informative snippets of the video, from which the final summary can be created. In this way, we are selecting key-shots/snippets as opposed to selecting key-frames. The first stage is to segment the temporal structure of the video. To do this, we use the KTS algorithm. KTS groups frames of similar content together \cite{Zhang2016}. Once we have these snippets of no more than 5 seconds each, we employ the same method used by \cite{Zhang2016} to rank each snippet based on their scores. Thus, we need to move from our frame-level scores to key-shots. We compute interval-level scores by averaging the individual scores of each frame within each interval to obtain these key-shots. Then, we rank all these intervals in descending order based on these interval scores. Lastly, we aim to select key intervals as the final summary such that they they less than a certain threshold. We choose this threshold to be 15\% of the original video duration, as done by \cite{Mahasseni2017, Zhang2016, Gygli2014, Song2015, Ji2019, Zhou2018}. This key interval selection is achieved using the 0/1 knapsack algorithm.

\subsubsection{Evaluation}
The model's performance can be evaluated using a common metric known as the F-score, which is defined as

\begin{equation}
\text{F-score} = 2 \times \frac{\text{precision} \times \text{recall}}{\text{precision} + \text{recall}} \times 100\%.
\end{equation}

Precision is the percentage of positive predictions that were correct, while recall is a measure of how well we find our all positive predictions. These are calculated as: 
\begin{equation}
\label{equation:Precision}
\text{Precision} = \frac{\textsl{Overlap duration of S and GT}}{\textsl{duration of S}},
\end{equation}
\begin{equation}
\label{equation:Recall}
\text{Recall} = \frac{\textsl{Overlap duration of S and GT}}{\textsl{duration of GT}},
\end{equation}
where S is the summary produced by our model and GT is the ground truth summary. In \eqnref{equation:Precision} and \eqnref{equation:Recall}, we calculate the temporal overlap between the models predictions and the ground truth.

For the evaluation, because both datasets have multiple user annotations, we pre-process the targets to be one single ground truth per video as done in previous works \cite{Zhang2016, Mahasseni2017, Ji2019, Zhou2018}. It is taken by thresholding by some value $\sigma$ to binarise the targets to be $0$ or $1$, depending on whether it is a summary frame or not. The probabilities produced by the model are then processed as well, whereby the $85^{\text{th}}$ percentile of values are taken from the model output. These values are then binarised. This means the only $15\%$ of the total output is considered to ensure that we end up with a realistic video summary.

\subsection{Baseline Model}
In this research, we first develop a baseline model from which we compare against to show the benefit of our proposed approach. This baseline model uses feature representations which we extract for each video frame. The model consists of 2 hidden layers with 256 units, both with sigmoid activation functions. The output layer is a single unit with a sigmoid activation for classification. The model uses a batch size of 8, binary cross-entropy as the loss function, and the Adam algorithm as the optimiser. We also sub-sample each video at 2fps, as mentioned previously.

\section{Experimental Results}
\label{sec:Experimental Results}

\subsection{Datasets}
There are numerous benchmark datasets that exist such as Thumb1k \cite{Liu2015}, TVSum50 \cite{Song2015} and SumMe \cite{Gygli2014} which can be used to evaluate video summarisation algorithms. These datasets have some additional information we might find useful, such as the meta-data. Even though the datasets are relatively new, they have previously been used as benchmark datasets for the problem \cite{Yuan2019, Zhang2016, Ji2019, Mahasseni2017, Zhou2018}, and as such, we use them to benchmark our results.

\subsubsection{SumMe Dataset}
SumMe dataset was originally introduced by \cite{Gygli2014}. It consists of 25 videos in both .mp4 and .webm formats. All videos differ in resolution and duration, ranging between 1-6 minutes. Scenes in these videos generally change slowly. Each video is accompanied by a .mat file which has some meta-data about the video. The additional information includes frames per second, number of frames in the video sequence, and video duration. Additionally each video has been manually annotated by 15-18 different users. That is, each user indicated snippets that they believed was worthy of being part of the summary. From this, the ground truth and user score for each frame was computed and also included as part of the dataset. The types of camera position of the videos are either moving, egocentric, or static, with the bulk of the videos taken with a moving camera.

\subsubsection{TvSum50 Dataset}
\label{subsec:TVSumDataset}
TvSum50 dataset was originally introduced by \cite{Song2015}. It consists of 50 videos all in .mp4 format. The videos differ in resolution and duration, ranging between 1-11 minutes. This dataset has 10 different categories, i.e. 5 videos for each category. The dataset is accompanied by two .tsv files, which include information about each video. In the first file, the video category, video name, video title, source URL, and video duration are included. In the second file, we are given 20 different people that annotated each video file. We are given the video name, video category, and importance score. The importance score ranges between 1-5, with 5 being very important, and 1 being unimportant. Users gave a single importance score for every 2 seconds of the video (i.e. every frame between the two importance scores will have the same value).

\subsection{Results}
\subsubsection{TvSum50 Dataset}

\begin{figure*}
	\centering
	\setlength{\tabcolsep}{1pt}
	\begin{tabular}{cc}
		\begin{tabular}[t]{cc}			
			\multicolumn{2}{c}{
				\includegraphics[width=0.40\linewidth, align=c]{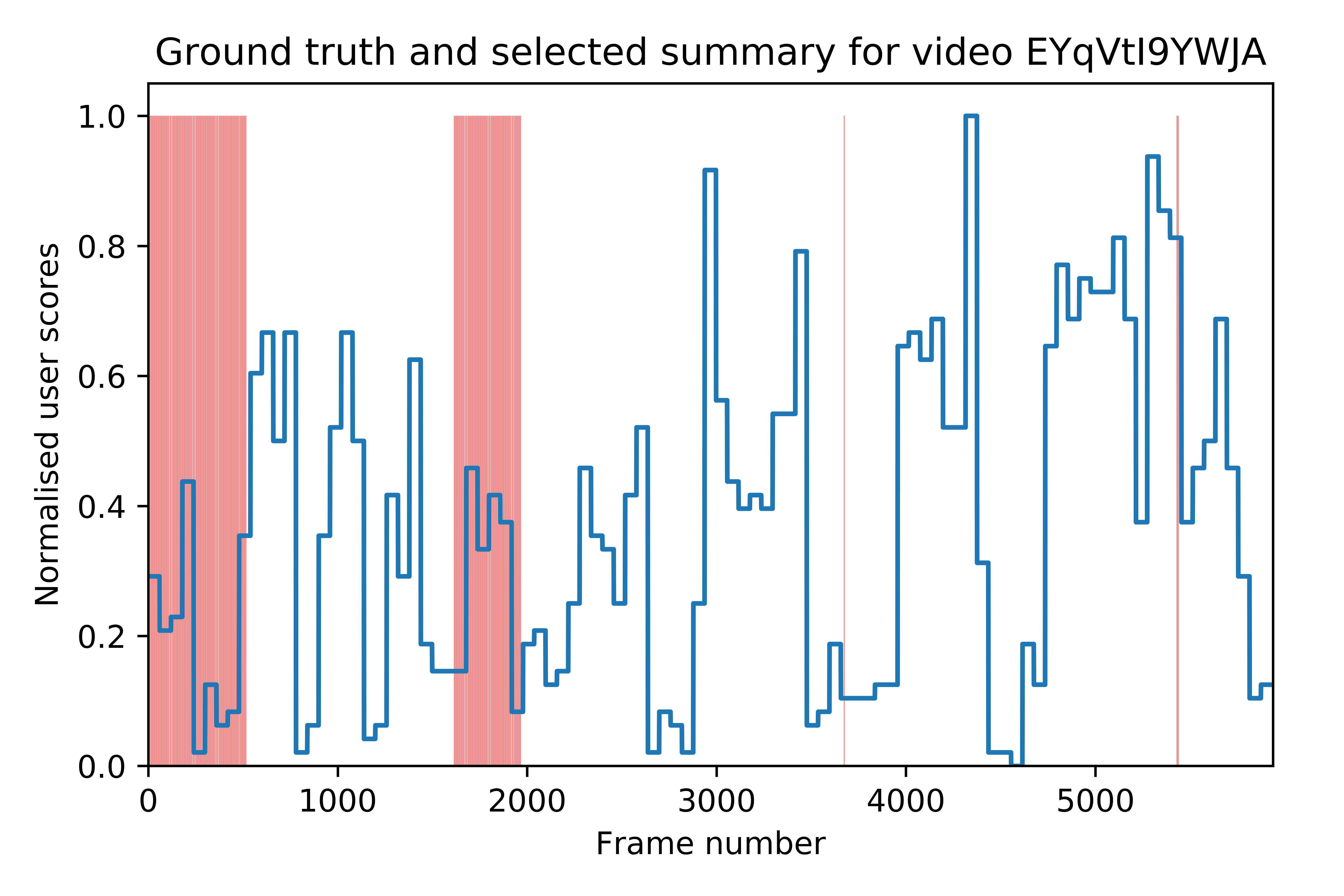}
			}
			
			\\
			
			\snip{0.24\linewidth}{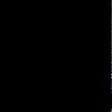}{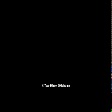}{First and last frames}
			&			
			\snip{0.24\linewidth}{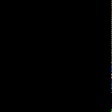}{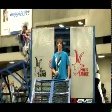}{Snippet 2--518} 
			\\
			
			\snip{0.24\linewidth}{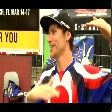}{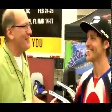}{Snippet 1614--1967} 
			&
			\snip{0.24\linewidth}{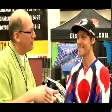}{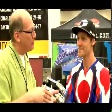}{Snippet 3673--3678} 
			\\
			
			\multicolumn{2}{c}{			
				\snip{0.24\linewidth}{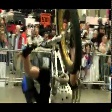}{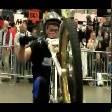}{Snippet 5429--5441} 
			}			
		\end{tabular}
		
		&

		\begin{tabular}[t]{cc}	
			
			\multicolumn{2}{c}{
				\includegraphics[width=0.40\linewidth, align=c]{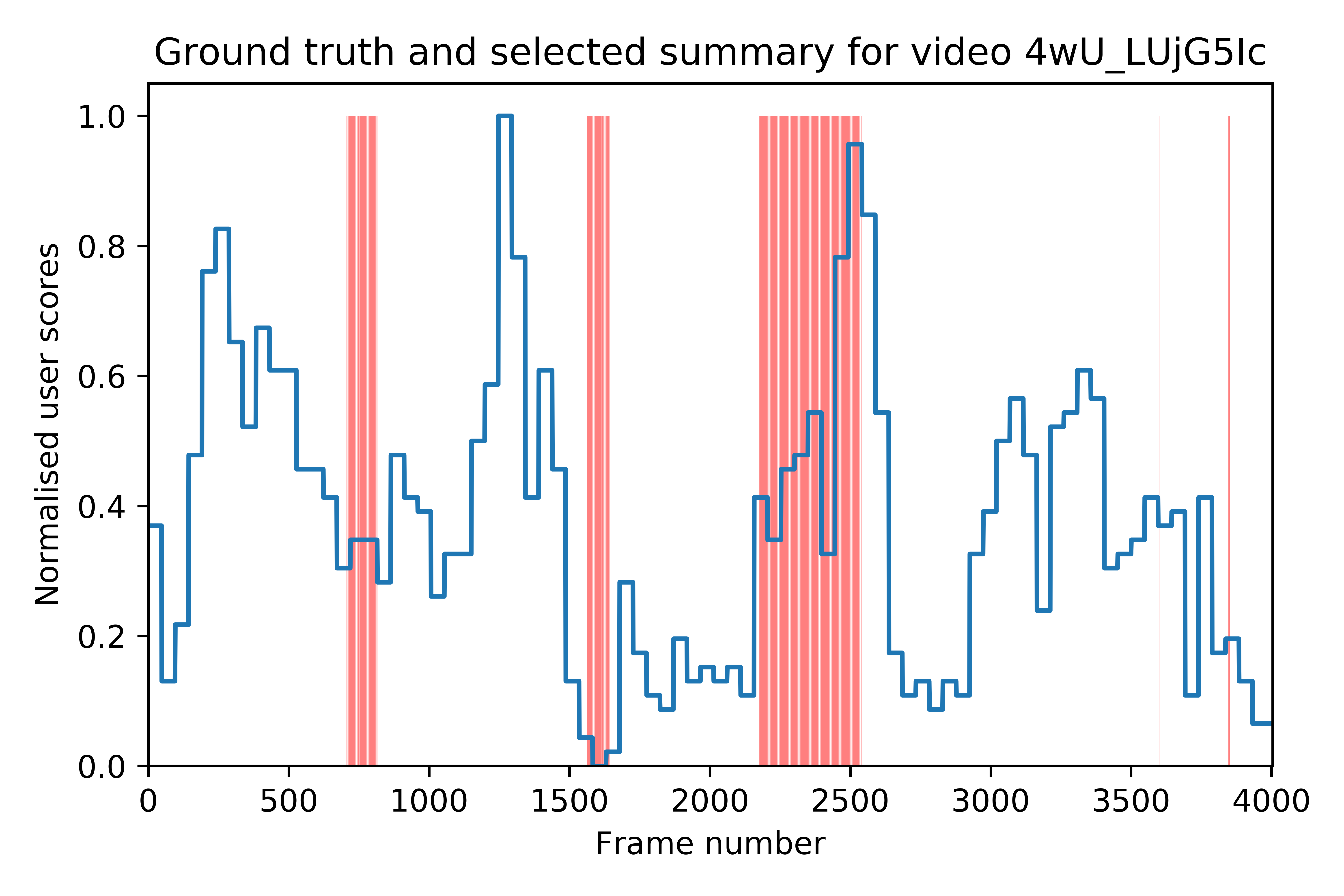}
			}
			
			\\
			
			\snip{0.24\linewidth}{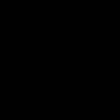}{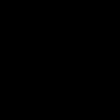}{First and last frames}
			&			
			\snip{0.24\linewidth}{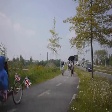}{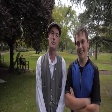}{Snippet 706--819} 
			\\
			
			\snip{0.24\linewidth}{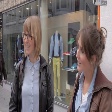}{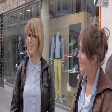}{Snippet 1564--1642} 
			&
			\snip{0.24\linewidth}{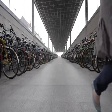}{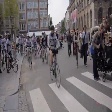}{Snippet 2174--2540} 
			\\
			
			\snip{0.24\linewidth}{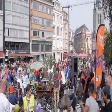}{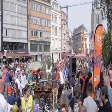}{Snippet 3599--3601} 
			&
			\snip{0.24\linewidth}{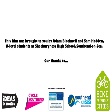}{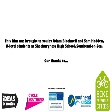}{Snippet 3847--3853}			
		\end{tabular} \\
		(a) Video ``EYqVtI9YWJA'' & (b) Video ``4wU\_LUjG5Ic''
	\end{tabular}
	
	\caption{Samples of bad (a) and good (b) summary results from TvSum50 dataset. The plots show the average user score per frame and the red rectangles signifies which frames the model chose as a summary. The images in each column are the first and last frames (in pair) of input video and snippets of summary selected by the model. The numbers for each snippet indicate the frame numbers of first and last frame of the corresponding snippet.}
	\label{fig:TvSum50_results}
\end{figure*}


\begin{table}
\centering
\caption{Results achieved by our SummaryNet model for different optical flow grid sizes. We choose to use a grid size of $15 \times 15$ for the remaining experiments in this paper. The values in this table correspond to the TvSum50 dataset.}
\begin{tabular}{|c|c|}
\hline
Grid Size & F-score (\%) \\ \hline \hline
$1 \times 1$ & 67.3 \\ \hline
$8 \times 8$ & 70.9 \\\hline 
$15 \times 15$ & 71.7 \\ \hline
$20 \times 20$ & 70.4 \\ \hline
\end{tabular}
\label{tab:OF_grid_size}
\end{table}

From \tableref{tab:OF_grid_size}, we see that grid size in the dense optical flow computation has a notable effect on accuracy. There is a trade-off between having too small a grid size (which would result in too much noise in the optical flow videos), and too large a grid size (which would result in not enough signal being present in the optical flow videos). We choose the grid size value that results in the best performance (i.e. $15 \times 15$) for all subsequent experiments.

\begin{table}
	\centering
	\setlength{\tabcolsep}{3pt}
	\caption{Comparison of video summarisation methods using F-score (\%) on TvSum50 dataset.  These results clearly show the advantage of using spatio-temporal features.}
	\begin{tabular}{|l|c|c|c|c|}
		\hline
		\textbf{Method} & \textbf{Features} & \textbf{RGB} & \textbf{Opt. Flow} & \textbf{RGB+Opt. Flow} \\ \hline \hline 
		Baseline & ResNet & - & - & 50.48 \\ \hline
		Baseline & I3D & 67.30 & 67.28 & 68.64 \\ \hline
		ConvNet & I3D & 70.00 & 67.40 & 68.40 \\ \hline
		ConvLSTM & I3D & 70.18 & 70.44 & 69.10 \\ \hline
		SummaryNet & I3D & 70.06 & 71.70 & \textbf{72.02} \\ \hline
	\end{tabular}
	\label{table:TvSum50_results}
\end{table}

The dataset consist of 50 videos. Evaluation is performed using a 5-fold cross-validation framework. As such, each testing set has at most 1 category, leaving 4 videos of the that same category for training. First, snippets of duration 16 frames are created for each video. For a video of length $N$ frames, we will have $N-16$ snippets (since we are using a stride of 1 frame). This process is repeated for both the RGB and the optical flow videos. These 16-frame snippets are then used to fine-tune the I3D networks. We then extract features from the second-last layer of the I3D network (the layer before the classification layer) as a feature representation for each frame. These features are then fed into the encoder-decoder model, and subsequently, the LSTM regression network. Since the TvSum50 dataset consists of 5 videos per category, we can expect relatively high results compared to that of the SumMe dataset, in which there are no categories, and instead just a random set of videos.

In \tableref{table:TvSum50_results}, we see the results of baseline-ResNet is more than 3\% lower than the results obtained by \cite{Zhang2016} who used GoogLeNet features with a similar model architecture. In this table we also clearly see the advantage of using spatio-temporal features. Since we are using a two-stream network, we get two separate predictions for a video. The results are combined and the average is taken as the final prediction. These scores are represented by the last column in the table. Our proposed method, SummaryNet is shown in the last row of \tableref{table:TvSum50_results}. This method clearly shows significant improvement in the overall results when compared to the baseline methods. This could possibly be due to the fact that we have 4 videos for each category used in training which makes it easier for the model to learn how to summarise a particular category. Another possibility is that most of the videos are slowly changing, this means that after performing the KTS algorithm to segment the video, the algorithm is able to distinguish between similar segments. Our approach performs better than baseline also because during the autoencoder stage, the most salient features are captured, i.e. the features that contribute the most to the make up of the representation feature. These features which are subsequently fed into the MLP stage clearly shows the power of using these salient features. Thus, by combining a unsupervised method with a supervised method, we are able to capture a low-dimensional feature space by using a under-complete structure, as well as allowing the algorithm to learn from human summaries to better mimic this task. 

In \tableref{table:TvSum50_results}, we see that I3D-based features serve as a better, more rich representation of the videos than ResNet features. This is evident from the respective accuracies of both approaches--with the baseline I3D approach achieving an accuracy of 68.64\%, compared to the 50.48\% from the ResNet baseline approach. This large difference can be attributed to the fact that the ResNet features only incorporate 2D spatial information into the representation, whereas the I3D features incorporate 3D spatial information (through the RGB stream), as well as explicitly modelling temporal information (through the optical flow stream). This shows that separately modelling spatial / appearance and temporal / motion information in videos is useful. The ConvNet I3D approach improves over the baseline I3D approach for both RGB and optical flow separately, however, when combined the performance is slightly worse. This is likely because in this particular model setup, the RGB and optical flow predictions sometimes strongly `disagree' on what frames are summary frames. As such, when averaging the results of both streams, the predictions become somewhat saturated. Another reason we attribute this slight decrease in performance of ConvNet I3D for the combined RGB and optical flow for this dataset to the way we are combining the output of the two streams. We use a simple averaging approach whereby the RGB and optical flow outputs of the sigmoid regression network are simply averaged. This simple scheme is likely a suboptimal way to fuse the information from the two streams and more complex information fusion methods can likely yield higher performance. The ConvLSTM I3D model improves upon the previous, standard ConvNet model on all metrics (RGB, optical flow, and combined). This is likely due to the bidirectional LSTM cells present in the regression sigmoid network, which are able to effectively model the temporal dynamics of the RGB and optical flow I3D representations. Lastly, the full SummaryNet model performs similarly to the ConvLSTM model on the separate RGB and optical flow streams. However, when combined there is a non-trivial increase in performance with the SummaryNet model. We attribute this performance increase to the encoder-decoder model, which is able to effectively capture the most salient information from the I3D representations, and remove any noise from the core signal that could potentially be detrimental in the sigmoid network's learning process. It is important to note that the KTS points are obtained by using the features that are supplied to the model which directly influences the systems results. Thus, the change points supplied by using the I3D features clearly demonstrate its superiority over using simple 2D spatial features, such as those from ResNet.

In \figureref{fig:TvSum50_results}(a), we see the frames selected by our model for the video name ``EYqVtI9YWJA'', where the model achieved a relatively low F-score of 31.69\%, and a total summary of $0.1499$. This refers to the summary being 14.99\% in length of the original video. This is under the video category \textit{BT} which translates to: attempting bike tricks. This video is taken with the camera moving and continuously alternates between the exact same interview state and a similar state where people are performing tricks on a motorbike. In the summary frames seen in \figureref{fig:TvSum50_results}(a), it is clear that similar frames were selected even though these frames appear multiple times over the video. We also see that the model selected the first few frames and then jumped to frames between 1614-1967. Thereafter, the model does not choose any other frames for the remainder of about 4000 frames. We would generally want to get summaries throughout the entire video, which will enable us to get a good synopsis of a video's content. This could be because of the large amount of movement/alternating camera views (going back and forth to similar views) that causes the model to perform poorly. Another reason could be the seemingly high disagreement between user scores for these types of videos makes it difficult for the model to learn a optimal summary. This again relates to the highly-subjective nature of the problem, lending to its inherent difficulty. One last potential reason for the poor performance in this case could be the fact that the I3D networks were fine-tuned on a relatively low spatial ($112 \times 112$) and temporal ($16$) resolution. As shown in \cite{Varol2018} for the task of action recognition, temporal resolution has a significant effect on accuracy, and spatial resolution has a somewhat lesser (but still noticeable) effect. Our video representations would thus likely have improved if we trained the I3D networks on snippets of higher spatial and temporal resolution. A higher temporal resolution could capture more of the temporal evolution of the videos, as well as more granular motion (and similarly for spatial resolution).

In \figureref{fig:TvSum50_results}(b), we see the frames selected by our model for the video name ``4wU\_LUjG5Ic'', where the model achieved a relatively high F-score of 97.91\%, and a total summary of $0.1426$. This is under the video category \textit{PR} which translates to: ``PaRade". This video is taken with the camera moving and continuously alternating between the different interview states and dissimilar states in which people are attending a parade. In the summary frames seen in \figureref{fig:TvSum50_results}(b), it is clear that frames of varying content were selected. It shows people on bicycles, followed by an interview, a different aspect of the parade, and finally an interview shot again. In the plot of \figureref{fig:TvSum50_results}(b), we see that shots are selected throughout the video which gives us a good synopsis of the video content. The model seemingly performs well when there are varying scenes in the video that are dissimilar from previous scenes. The low resolution of the snippets for the I3D networks does not seem to be as much of an issue for these types of videos, most likely because the various scenes are very different, and easily capture using a relatively small snippet length.

\subsubsection{SumMe Dataset}

\begin{figure*}
	\centering
	\setlength{\tabcolsep}{1pt}
	\begin{tabular}{cc}
		\begin{tabular}[t]{cc}			
			\multicolumn{2}{c}{
				\includegraphics[width=0.40\linewidth, align=c]{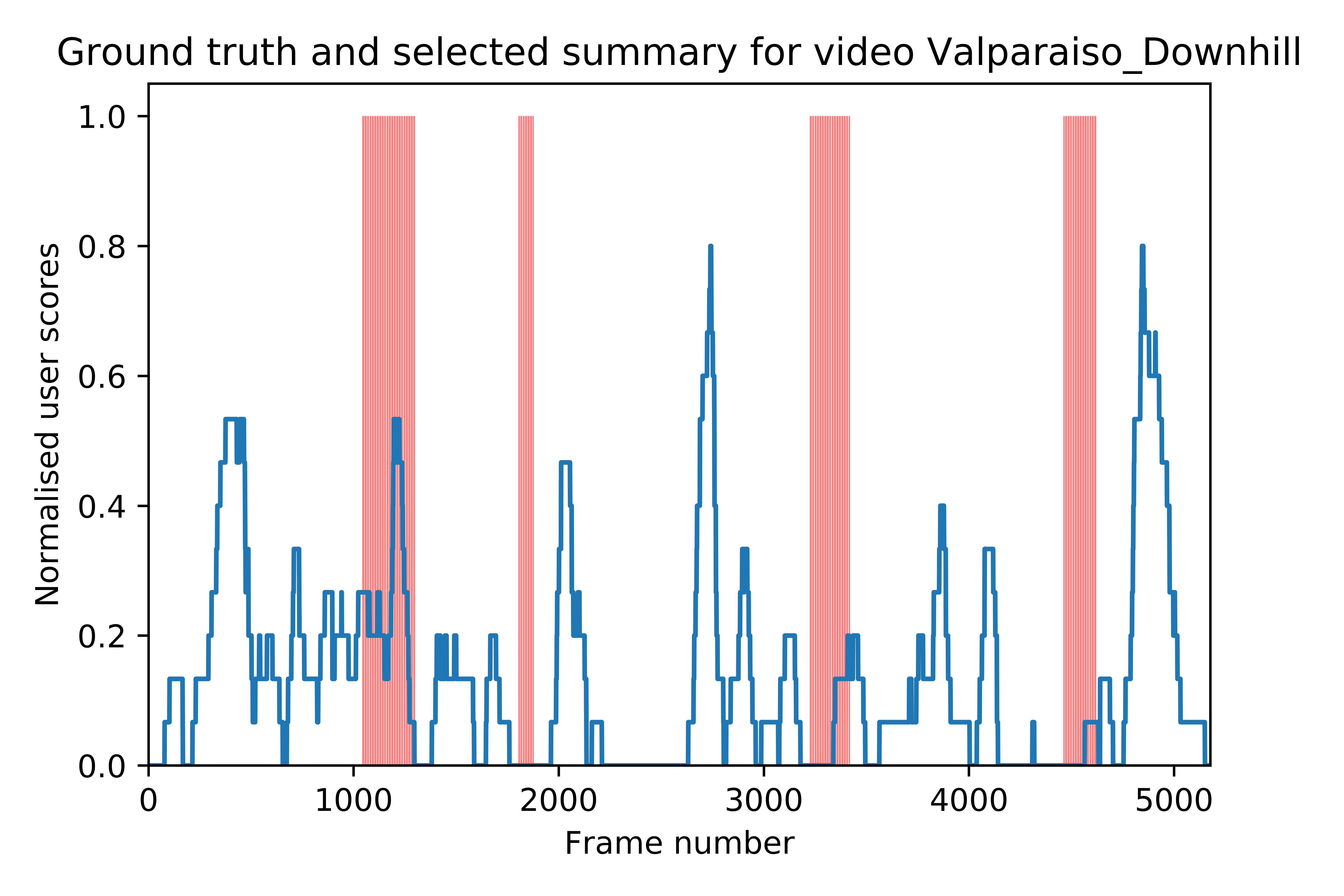}
			}
			
			\\
			
			\snip{0.24\linewidth}{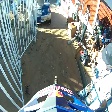}{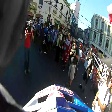}{First and last frames}
			&			
			\snip{0.24\linewidth}{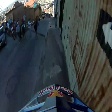}{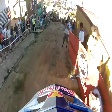}{Snippet 1043--1299} 
			\\
			
			\snip{0.24\linewidth}{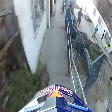}{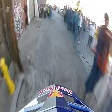}{Snippet 1804--1878} 
			&
			\snip{0.24\linewidth}{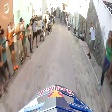}{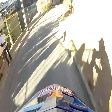}{Snippet 3223--3420} 
			\\
			
			\multicolumn{2}{c}{			
				\snip{0.24\linewidth}{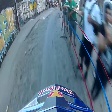}{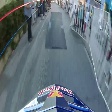}{Snippet 4461--4621} 
			}			
		\end{tabular}
		
		&

		\begin{tabular}[t]{cc}	
			
			\multicolumn{2}{c}{
				\includegraphics[width=0.40\linewidth, align=c]{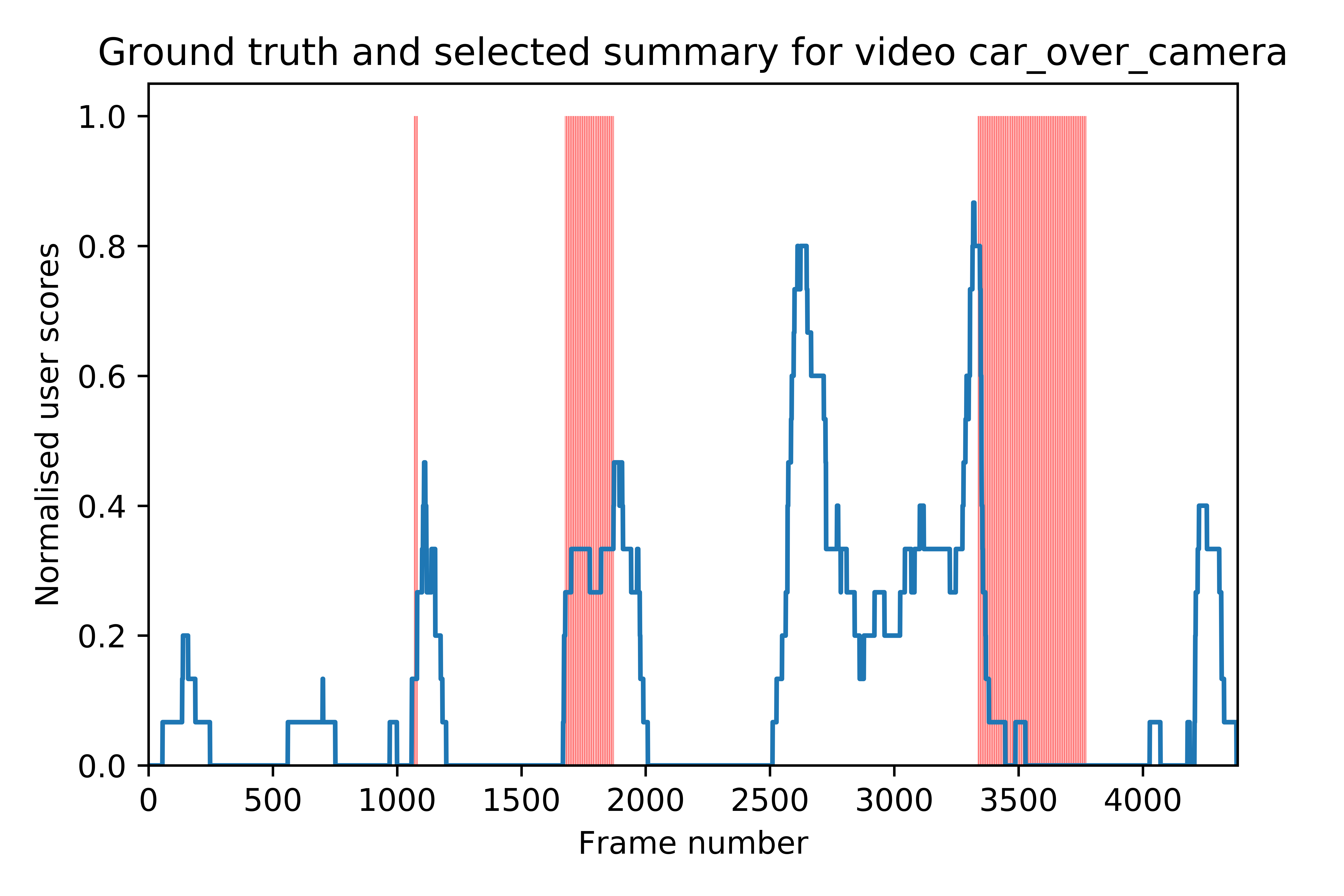}
			}
			
			\\
			
			\snip{0.24\linewidth}{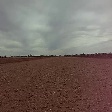}{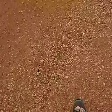}{First and last frames}
			&			
			\snip{0.24\linewidth}{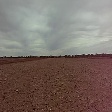}{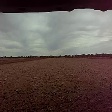}{Snippet 1068--1083} 
			\\
			
			\snip{0.24\linewidth}{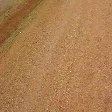}{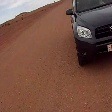}{Snippet 1675--1872} 
			&
			\snip{0.24\linewidth}{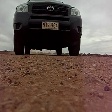}{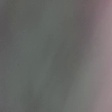}{Snippet 3336--3773} \\
			
		\end{tabular} \\
		(a) Video ``Valparaiso\_Downhill'' & (b) Video ``car\_over\_camera''
	\end{tabular}
	
	\caption{Samples of bad (a) and good (b) summary results from SumMe dataset. The plots show the average user score per frame and the red rectangles signifies which frames the model chose as a summary. The images in each column are the first and last frames (in pair) of input video and snippets of summary selected by the model. The numbers for each snippet indicate the frame numbers of first and last frame of the corresponding snippet.}
	\label{fig:SumMe_results}
\end{figure*}

\begin{table}
	\centering
	\setlength{\tabcolsep}{3pt}
		\caption{Comparison of video summarisation methods using F-score (\%) on SumMe dataset.  These results clearly show the advantage of using spatio-temporal features.}
	\begin{tabular}{|l|c|c|c|c|}
		\hline
		\textbf{Method} & \textbf{Features} & \textbf{RGB} & \textbf{Opt. Flow} & \textbf{RGB+Opt. Flow} \\ \hline \hline
		Baseline & ResNet  & - & - & 38.06 \\ \hline
		Baseline & I3D  & 33.86 & 41.45 & 40.64 \\ \hline
		ConvNet & I3D  & 33.90 & 41.77 & 42.85 \\ \hline
		ConvLSTM & I3D  & 33.88 & 42.58 & 43.75 \\ \hline
		SummaryNet & I3D  & 35.83 & 43.07 & \textbf{44.60} \\ \hline
	\end{tabular}
	\label{table:SumMe_results}
\end{table}

This dataset consists of 25 videos, with 5 videos random chosen for testing. The bulk of the videos in this dataset are taken while the camera is moving. The rest consists of egocentric and static videos, which have 4 videos each. Features for each video frame are extracted in the same way described above. It should be noted that the fine-tuning of I3D networks is performed specifically for the current dataset and not both. Since the videos in the SumMe dataset are not arranged by category, and are somewhat arbitrary, we can expect lower results as compared to the TvSum50 dataset. Another reason why the SumMe dataset is challenging is because even though the videos are slowly changing - almost 70\% of the videos supplied are videos where the camera is moving, which makes it difficult to gather change points from the KTS algorithm. Future work could possibly involve finding alternative algorithms to address these types of videos.

In \tableref{table:SumMe_results}, we see a similar trend to that of \tableref{table:TvSum50_results}. The baseline I3D model outperforms the baseline ResNet model, due to the incorporation of 3D spatial and temporal information into the representations. The trends follows that of the TvSum50 dataset for the other I3D-based models as well, with the combined SummaryNet model performing best. However, it should be noted that the performance gains for each model is not as drastic as the TvSum50 dataset (e.g. baseline ResNet vs baseline I3D), and performance is on average much lower. The dataset does not consist of well-defined classes with multiple samples each (as in TvSum50), but instead consists of a random set of videos. Additionally, the videos in SumMe consists of highly non-static backgrounds, with lots of noise and erroneous motion that would affect the modelling process, and the video representations.

In \figureref{fig:SumMe_results}(a), we see the frames selected by our model for the video name ``Valparaiso\_Downhill'', where the model achieved a relatively low F-score of 23.79\%, and a total summary of $0.1443$. This video is taken with the camera moving (egocentric) and continuously alternates between different scenes i.e. fast pace scene changes. In the summary frames seen in \figureref{fig:SumMe_results}(a), the selected frames seem very similar, even though they are taken at different points in the video. We also see that the model selected 4 different shots throughout the video, which is ideally what we would want. However, the last shot chosen by our model shown in \figureref{fig:SumMe_results}(a) selected frames between 4461-4621 which has almost a 0 average user score. Similar to what was discussed for the TvSum50 dataset where the model performed poorly, this could be because of the large amount of movement/alternating camera views (going back and forth to similar views) that causes the model to perform poorly. The optical flow estimates for these types of videos would be very noisy, resulting in less effective video representations.

In \figureref{fig:SumMe_results}(b), we see the frames selected by our model for the video name ``car\_over\_camera'', where the model achieved a relatively high F-score of 50.39\%, and a total summary of $0.1162$. This video is taken with the camera mostly stationary and focus on a object in the view of the camera. In this video, the scene changes slowly and differences in the scenes are clearly visible. In the summary frames seen in \figureref{fig:SumMe_results}(b), it is clear that frames of varying content were selected. The camera is placed on the ground (sand) and the car attempts to drive over it. In the plot of \figureref{fig:SumMe_results}(b), we see that shots are selected where there was high agreement between user scores. Although it missed the highest peaks between frames 2500-3500, it still gives a good synopsis, and achieves a relatively high F-score. Since there are varying scenes that are slower changing the model seemingly performs well when there are scenes that are dissimilar from previous scenes. This observation mirrors that of the TvSum50 dataset performance.

\subsection{Comparing Video Summarisation Methods}

\begin{table}
	\centering
	\caption{Comparison of video summarisation methods using F-score (\%). The best results are shown in dark blue and second best is shown in light blue.}
	\begin{tabular}{|l|c|c|}
		\hline
		\textbf{Method} & \textbf{SumMe} & \textbf{TvSum50} \\ \hline \hline
		\cite{Gygli2014} & 39.7 & - \\ \hline
		\cite{Song2015} & - & 51.3 \\ \hline
		\cite{Zhang2016} & 38.6 & 54.7 \\ \hline
		\cite{Mahasseni2017} & 41.7 & 56.3 \\ \hline
		\cite{Zhou2018} & 42.1 & 58.1 \\ \hline
		\cite{Rochan2018} & \cellcolor{blue!10}\textbf{47.5} & 56.8 \\ \hline
		\cite{Yuan2019} & - & 57 \\ \hline
		\cite{Ji2019} & 44.4 & \cellcolor{blue!10}\textbf{61} \\ \hline
		\cite{Elfeki2019} & 40.1 & 56.3 \\ \hline
		\cite{CycleNet} & 41.9 & 57.6 \\ \hline
		\cite{CSNet} & \cellcolor{blue!25}\textbf{51.3} & 58.8 \\ \hline
		SummaryNet (ours) & \textbf{44.6} & \cellcolor{blue!25}\textbf{72.02} \\ \hline
	\end{tabular}
	\label{table:ComparingResults}
\end{table}

In \tableref{table:ComparingResults}, we compare video summarisation methods.  All the methods shown in this table use deep learning, often leveraging deep spatial features (e.g. GoogLeNet features). For the TvSum50 dataset, SummaryNet outperforms the state-of-the-art \cite{Ji2019}  by more than 11\%. This significant increase in accuracy is attractive since it demonstrates the power of SummaryNet. For the SumMe dataset we achieve competitive results, where the state-of-the-art \cite{CSNet} is around 6.7\% better than SummaryNet. However, SummaryNet outperforms \cite{CSNet} on TvSum50 dataset by a large margin of more than 13\%. This could be due to the fact that the SumMe dataset is more challenging dataset compared to the TvSum50 dataset since the SumMe dataset has no categorical structure associated with it. Additionally, the previously-discussed limited temporal and spatial resolutions of our snippets for fine-tuning I3D likely resulted in poorer video representations. We leave it as future work to train I3D on higher resolutions snippets. It should be noted, however, that this is a huge computational challenge. Overall, SummaryNet achieves the state-of-the-art performance on TvSum50 dataset, and is competitive with state-of-the-art for the SumMe dataset.

Our approach is able to capture the most salient features for each frame as well as exploit temporal dependencies which makes our method perform well on the TvSum50 and SumMe datasets. The temporal information seemingly adds invaluable information to the overall system accuracy. One important aspect of our method that should be addressed in future work is the need for a better way to combine the RGB and optical flow streams. This is made clear by the tables results mentioned above. However, the overall system accuracy still benefits from using both the spatial and temporal aspects present in video data.

\section{Conclusion}
\label{sec:Conclusion}

SummaryNet employs both spatial and temporal features extracted with a state-of-the-art 3D CNN architecture, and proves to be robust on both the SumMe and TvSum50 datasets. It achieves competitive results on the SumMe dataset and achieves state-of-the-art on the TvSum50 dataset.

The model we designed does well on videos that have varying scenes throughout the video. This suggests the need to develop a method that can deal with videos that have similar scenes throughout the video and still produce a relatively good synopsis of the video. Along with this, future models can include attention mechanisms to focus on certain generic scenes in a video. Additionally, one can also consider adding acoustic or audio cues, and other forms of video meta-data (as separate modelling streams) to aid the task of video summarisation to push performance to near human levels.

\bibliographystyle{unsrt}  
\bibliography{references}  






\end{document}